\documentclass[runningheads]{llncs}
\pdfoutput=1
\usepackage{graphicx}
\usepackage{amsmath}
\usepackage[misc]{ifsym}
\usepackage{amssymb}
\begin{document}
\title{A Medical Semantic-Assisted Transformer for Radiographic Report Generation}
\author{Zhanyu Wang\inst{1\ast} \and
Mingkang Tang\inst{2\ast} \and
Lei Wang\inst{3} \and
Xiu Li\inst{2} \and
Luping Zhou\inst{1(\textrm{\Letter})}}
\authorrunning{F. Author et al.}
%
\institute{University of Sydney, NSW, Australia\\
\email{\{zhanyu.wang, luping.zhou\}@sydney.edu.au} \and
Shenzhen International Graduate School of Tsinghua University, Shenzhen, China
\email{tmk20@mails.tsinghua.edu.cn} \quad
\email{li.xiu@sz.tsinghua.edu.cn} \and
University of Wollongong, NSW, Australia\\
\email{leiw@uow.edu.au}}
%
%
%

%
\maketitle              
\renewcommand{\thefootnote}{}
\footnotetext{\inst{\ast}Equal contribution}
\footnotetext{\inst{\textrm{\Letter}}Corresponding Author}
\renewcommand{\thefootnote}{\arabic{footnote}}
\setcounter{footnote}{0}
\begin{abstract}
Automated radiographic report generation is a challenging cross-domain task that aims to automatically generate accurate and semantic-coherence reports to describe medical images. Despite the recent progress in this field, there are still many challenges  at least in the following aspects. First, radiographic images are very similar to each other, and thus it is difficult to capture the fine-grained visual differences using CNN as the visual feature extractor like many existing methods. Further, semantic information has been widely applied to boost the performance of generation tasks (e.g. image captioning), but existing methods often fail to provide effective medical semantic features. Toward solving those problems, in this paper, we propose a memory-augmented sparse attention block utilizing bilinear pooling to capture the higher-order interactions between the input fine-grained image features while producing sparse attention. Moreover, we introduce a novel Medical Concepts Generation Network (MCGN) to predict fine-grained semantic concepts and incorporate them into the report generation process as guidance. Our proposed method shows promising performance on the recently released largest benchmark MIMIC-CXR. It outperforms multiple state-of-the-art methods in image captioning and medical report generation.

\end{abstract}
%
%

\section{Introduction}~\label{introduction}
Automated radiographic report generation aims to generate a paragraph to address the observations and findings of a given radiology image. It has significant application scenarios in reducing the workload and mitigating the diagnostic errors of radiologists who are under pressure to report increasingly complex studies in less time, especially in emergency radiology reporting.
Due to its clinical importance, medical report generation has gained increasing attention. The encoder-decoder paradigm has prevailed in this field, motivated by its success in generic image captioning~\cite{2015DeepVisualSemantic,vinyals2015showandtell,xu2016showattandtell,2016semanticattention,2017Knowing,topdown,2017Ontheautomatic,2020Meshed,chen2020generating,CVPR201_PPKD}, where many approaches apply similar structures including an encoder based on convolutional neural networks (CNN), and a decoder based on recurrent neural networks (RNN)~\cite{2015DeepVisualSemantic,vinyals2015showandtell,xu2016showattandtell}. Furthermore, several research works employ carefully-designed attention mechanism to boost performance~\cite{2017Knowing,topdown,2017Ontheautomatic}. Most recently, there have been efforts starting to explore transformer-structured framework~\cite{2020Meshed,chen2020generating,CVPR201_PPKD} in this field.

Despite these progresses, there are still some non-negligible problems in the medical report generation methods rooted in image captioning due to the non-trivial different characteristics of these two tasks. First, unlike natural images, radiographic images are very similar to each other. The medical report generation methods using the CNN-based image encoder extract image-level features due to the lack of image region annotations in such applications, which could not well cater for the local details reflecting the fine-grained image patterns of clinic importance. This problem could be somewhat mitigated by employing the recently developed Vision Transformer model~\cite{dosovitskiy2021image} that explores the dependency of image regions without generating regional proposals. However, there still lacks mechanism to better treat fine-grained patterns.
Second, the differences between medical reports are also fine-grained, that is, the reports are dominated by similar sentences describing the common content of the images while the disease-related words may be submerged. A possible remedy is to incorporate  semantic textual concepts into the model training to guide the report generation. 
However, such information has not been well explored for medical report generation, especially in the recently popular transformer-structured models.

To cope with the limitations mentioned above, we propose the following solutions to advance radiographic report generation. Firstly, we carefully deal with the fine-grained differences existed in radiographic images from two aspects. On the one hand, we utilize CLIP~\cite{clip} rather than ResNet as our visual feature extractor. The advantage of CLIP is two-fold: i) it is built upon Vision Transformer that extracts regional visual features and relationships, and ii) it is trained by matching image-text pairs and thus produces text-enhanced image features. On the other hand, more importantly, we introduce an memory-augmented sparse attention for high-order feature interactions and embed it into the transformer encoders and decoders. This attention makes use of bilinear pooling that proves to be effective for fine-grained visual recognition~\cite{lin2015bilinear}, memorizes historical information for long report generation, and produces sparse attention for efficient report generation. 
Second, to incorporate semantic textual concepts, we introduce a novel lightweight medical concepts generation network to predict fine-grained semantic concepts and incorporate them into medical report generation process, which is different from the usage of sparse medical tags in the existing methods~\cite{2017Ontheautomatic}. In sum, our main contributions include: 1) We introduce a bilinear-pooling-assisted sparse attention block and embed it into a transformer network to capture the fine-grained visual difference existed between radiographic images; 2) We propose a medical concepts generation network to provide enriched semantic information to benefit radiographic report generation; 3) We extensively validate our model on the recently released largest dataset MIMIC-CXR. The results indicate that our framework outperforms multiple state-of-the-art methods in image captioning and medical report generation.
\section{Methodology}
The proposed framework adopts the encoder-decoder paradigm, where the transformer encoder and decoder are embedded with a Memory-augmented Sparse Attention block (MSA) to capture the higher-order interactions between the input fine-grained image features. Meanwhile, a Medical semantic Concepts Generation Network (MCGN) is incorporated into the report generation model to further improve the performance. In the following, we first introduce the MSA block in Section~\ref{sec:2-1} and how the MSA block are embedded into the encoder and decoder in Section~\ref{sec:2-2} and Section~\ref{sec:2-4}, and our proposed MCGN in Section~\ref{sec:2-3}.


\begin{figure}[t]
\includegraphics[width=\textwidth]{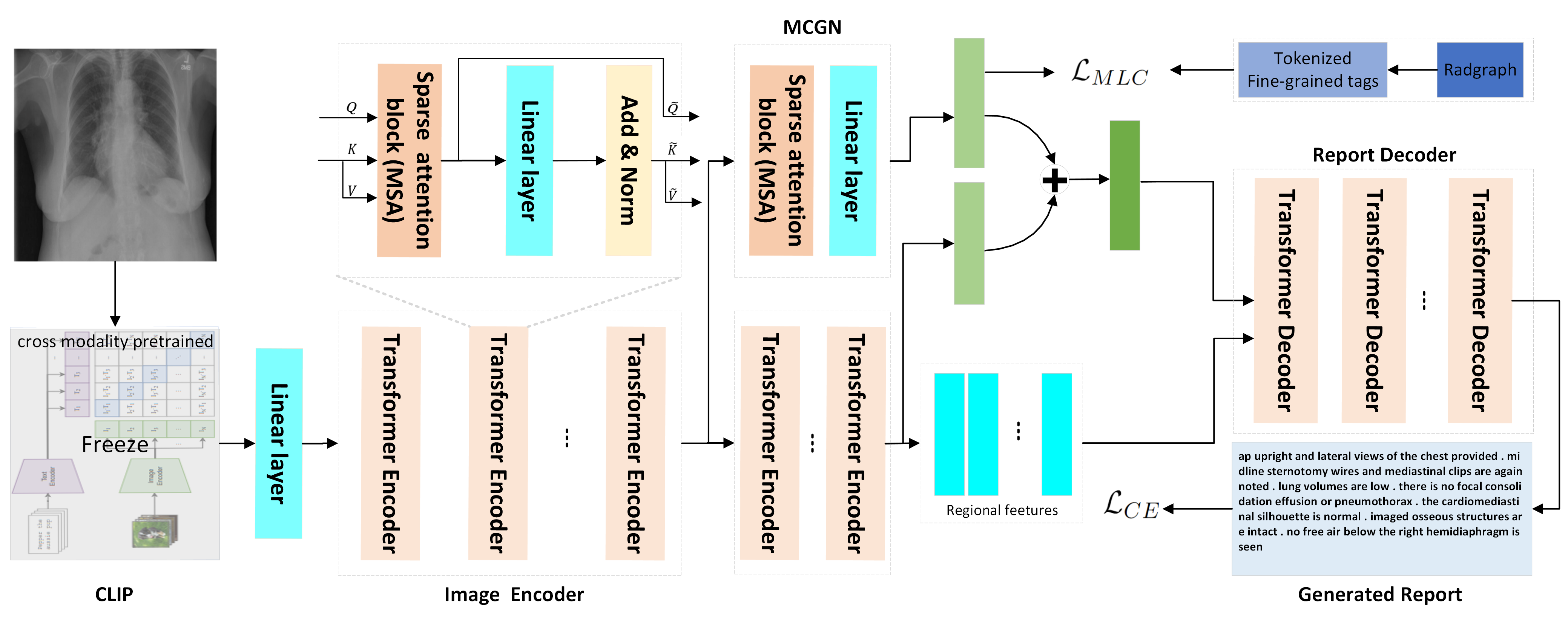}
\caption{An overview of the proposed framework, which comprises an Image Encoder, a Medical Concepts Generation Network (MCGN) and a Report Decoder. The transformer encoder and decoder are embedded with a Memory-augmented Sparse nonlinear Attention (MSA) to capture the higher-order interactions between the input fine-grained image features. Meanwhile, we inject a medical concepts generator to provide semantic information to facilitate report generation.} 
\label{fig:1}
\end{figure}


\subsection{Memory-augmented Sparse Attention of High-order Interaction}\label{sec:2-1}
Given the three input matrices, queries $\mathbf{Q} \in \mathbf{R}^{\mathbf{N} \times \mathbf{D}_q}$, keys $\mathbf{K} \in \mathbf{R}^{\mathbf{N} \times \mathbf{D}_k}$ and values $\mathbf{V} \in \mathbf{R}^{\mathbf{N} \times \mathbf{D}_v}$, the conventional self-attention block~\cite{vaswani2017attentionisallyouneed} employed by transformer first computes the dot products (element-wise sum) of each query with all keys and then applies a softmax function to obtain the weights on the values. Formally, the attention is calculated by: $\mathrm{Attention}(\mathbf{Q},\mathbf{K},\mathbf{V}) = \mathrm{softmax}(\frac{\mathbf{QK}^\mathbf{T}}{\sqrt{\mathbf{d}_k}})\mathbf{V}$. It exploits only $1^{st}$ order interaction of the input vectors since the weight is derived from the linear fusion of the given query and keys via element-wise sum. However, the linear attention may not be sufficient to capture the fine-grained visual differences among the input radiographic images. Inspired by the recent success of bilinear pooling applied in fine-grained visual recognition, we inject bilinear-pooling into the self-attention to capture the $2^{nd}$ or even higher-order (by stacking these blocks) interactions of the input fine-grained visual features.

Specifically, we assume the global feature as the query $\mathbf{Q} \in \mathbf{R}^{\mathbf{D}_q}$, the regional features as the key $\mathbf{K} \in \mathbf{R}^{\mathbf{N} \times \mathbf{D}_k}$ and value $\mathbf{V} \in \mathbf{R}^{\mathbf{N} \times \mathbf{D}_v}$. To record the historical information, we extent the set of keys and values with additional "memory-slots" to encode and collect the features from all the previous processes. The key and value of our memory-augment attention can be defined as: $\hat{\mathbf{K}} = [\mathbf{K}, \mathbf{M}_k]$ and $\hat{\mathbf{V}} = [\mathbf{V}, \mathbf{M}_v]$, respectively, where $\mathbf{M}_k$ and $\mathbf{M}_v$ are learnable matrices with $n_m$ rows, and $[\cdot, \cdot]$ indicates concatenation. Then, a low-rank bilinear pooling~\cite{lowrank} is performed to obtain the joint bilinear query-key $\mathbf{B}_k$ and query-value $\mathbf{B}_v$ by 
\begin{equation}\label{lowrank}
    \mathbf{B}_k = \sigma(\mathbf{W}_k \hat{\mathbf{K}}) \odot \sigma(\mathbf{W}_q^k\mathbf{Q}),\quad
    \mathbf{B}_v = \sigma(\mathbf{W}_v \hat{\mathbf{V}}) \odot \sigma(\mathbf{W}_q^v\mathbf{Q})
\end{equation}
where $\mathbf{W}_k \in \mathbf{R}^{\mathbf{D}_B \times \mathbf{D}_k}$, $\mathbf{W}_v \in \mathbf{R}^{\mathbf{D}_B \times \mathbf{D}_v}$, $\mathbf{W}_q^k \in \mathbf{R}^{\mathbf{D}_B \times \mathbf{D}_q}$ and $\mathbf{W}_q^v \in \mathbf{R}^{\mathbf{D}_B \times \mathbf{D}_v}$ are learning parameters, $\sigma$ denotes ReLU unit, and $\odot$ represents element-wise multiplication. Next, we use a linear layer to project ${\mathbf{B}_k} \in \mathbf{R}^{\mathbf{D}_c \times \mathbf{D}_B}$ into a intermediate representation $\acute{\mathbf{B}_k} = \sigma(\mathbf{W}_B^k \mathbf{B}_k)$, then use another linear layer to map $\acute{\mathbf{B}_k}$ from $\mathbf{D}_c$ dimension to 1 dimension to obtain the spatial-wise attention weight $\alpha_s \in \mathbb{R}^{\mathbf{D}_c \times 1}$. Unlike~\cite{pan2020xlinear} using softmax to normalize $\alpha_s$, we utilized another ReLU unit to prune out all negative scores of low query-key relevance, automatically ensuring the sparse property of the attention weight $\beta_s = \sigma(\alpha_s)$. Meanwhile, we perform a squeeze-excitation operation~\cite{hu2018squeeze} to $\acute{\mathbf{B}_k}$ to obtain channel-wise attention $\beta_c = \mathrm{sigmoid}(\mathbf{W}_c)\bar{\mathbf{B}}$, where $\mathbf{W}_c \in \mathbf{R}^{\mathbf{D}_B \times \mathbf{D}_c}$ is learnable parameters and $\bar{\mathbf{B}} \in \mathbf{R}^{\mathbf{D}_B \times 1}$ is an average pooling of $\acute{\mathbf{B}_k}$. The output attended features of our memory-augmented sparse attention integrate the enhanced bilinear values with spatial and channel-wise bilinear attention $\hat{\mathbf{Q}} = \mathrm{Attention}(\mathbf{K}, \mathbf{V}, \mathbf{Q})  = \beta_c \odot \mathrm{LN}(\beta_s\mathbf{B}_v)$. where $\mathrm{LN}(\cdot)$ denotes variants of layer normalization~\cite{ba2016layer}. 

Such structure design benefits the model in three ways. First, it can explore higher-order interactions between the input single-model (in the encoder) or multi-model (in the decoder) features, resulting in a more robust representative capacity of the output attended features. Second, the memory tokens we embed into the MSA block can record the previous generation process and collect historical information, which can be valuable for our task's long report generation. Third, we propose a softmax-free sparse attention module, pruning out all negative attention scores and producing sparse attention of transformer decoder with higher efficiency in the process of report generation. We measured the speedup per training step on 500 steps with about 16 samples per batch to compare the running efficiency of softmax-base attention and our sparse Relu-based attention. We perform three runs on a single NVIDIA TESLA V100 and report average results. Without performance degradation, our sparse attention can reduce the training time from 0.497s per batch to 0.488s per batch.

\subsection{Image Encoder}~\label{sec:2-2}
\noindent Let’s denote an input image by $\mathbf{I}$. The pre-trained CLIP model (ViT-B/16) is utilized to extract the regional features of I: $f = \mathrm{CLIP}(\mathbf{I})$, where $f \in \mathbb{R}^{D_c \times D_f}$. We take the mean-pooled feature embedding $f$ as the initial $\mathbf{Q}^{(0)} = \frac{1}{N_c}\sum_{i=1}^{N_c}f_i$, and $f$ as the initial $\mathbf{K}^{(0)}$ and $\mathbf{V}^{(0)}$, and feed them into the encoding layers.

\noindent\textbf{Encoding Layer}  We embed our memory-augmented sparse attention into a Transformer-like layer. Formally, for the m-th transformer encoder $TE_m$, we take the previous output attended feature $\hat{\mathbf{Q}}^{(m-1)}$ as the input query, couple it with the current input keys $\mathbf{K}^{m-1} = \left\{ \mathbf{k}_i^{(m-1)} \right\}_{i=1}^N$, and values $\mathbf{V}^{(m-1)} = \left\{ \mathbf{v}_i^{m-1} \right\}_{i=1}^N$, which could be expressed as $ \mathbf{K}^{(m)}, \mathbf{V}^{(m)}, \mathbf{Q}^{(m)} = \mathrm{TE}_m(\mathbf{K}^{(m-1)}, \mathbf{V}^{(m-1)}, \hat{\mathbf{Q}}^{(m-1)})$, with $\hat{\mathbf{Q}}^{(m)} = \mathrm{Attention}(\mathbf{K}^{(m-1)}, \mathbf{V}^{(m-1)}, \hat{\mathbf{Q}}^{(m-1)})$ computed by MSA block( see sec.~\ref{sec:2-1}).
Then, all the keys and values are further updated conditioned on the new attended feature $\hat{\mathbf{Q}}^{(m)}$ by equation: $\mathbf{k}_i^{(m)} = \mathrm{LN}(\sigma(\mathbf{W}_m^k[\hat{\mathbf{Q}}^{(m)}, \mathbf{k}_i^{(m-1)}]) + \mathbf{k}_i^{(m-1)})$ and $\mathbf{v}_i^{(m)} = \mathrm{LN}(\sigma(\mathbf{W}_m^v[\hat{\mathbf{Q}}^{(m)}, \mathbf{v}_i^{(m-1)}]) + \mathbf{v}_i^{(m-1)})$, respectively,
where $\mathbf{W}_m^k$ and $\mathbf{W}_m^v$ are learnable parameters. $[,]$ indicates concatenation. This means that each key or value is concatenated with the new attended feature, followed with a residual connection and layer normalization~\cite{ba2016layer}. Noted that the entire image encoder consists of $M=6$ encoding layers in this paper. In particular, when $m = 1$, $\hat{\mathbf{Q}}^{(0)} = \mathbf{Q}^{(0)} = \frac{1}{\mathbf{N}_c}\sum_{i=1}^{\mathbf{N}_c}f_i$.
\subsection{Medical Concepts Generation Network}~\label{sec:2-3}
In order to train the Medical Concepts Generation Network (MCGN), we utilize RadGraph~\cite{jain2021radgraph} to extract the pseudo-medical concepts as the ground-truth for multi-label classification (MLC). Specifically, the RadGraph is a knowledge graph of clinic radiology entitles and relations based on full-text chest x-ray radiology reports. We use these entities as the medical semantic concepts and finally select 768 concepts according to their occurrence frequency. 

Our MCGN consists of an MSA block for processing the intermediate features generated by the m-th Encoding layer of Image Encoder. The output of MSA is $\mathbf{V}_c = \mathrm{Attention}(\mathbf{K}^{(m-1)}, \mathbf{V}^{(m-1)}, \hat{\mathbf{Q}}^{(m-1)})$, where $m \in [1, M]$ denotes the m-th encoding layer. Then, a linear projection maps $\mathbf{V}_c$ from $D$ to $K$ dimensions, where $K$ is the number of medical concepts. The overall medical concepts classification loss can be expressed as:
\begin{equation}
\begin{aligned}
        {\mathcal L}_{MLC} = - \frac{1}{K} \cdot \sum_i y_i \cdot \log((1 + \exp(-x_i))^{-1}) \\
                         + (1-y_i) \cdot \log\left(\frac{\exp(-x_i)}{1 + \exp(-x_i)}\right),
\end{aligned}
\label{equ:l_mlc}
\end{equation}
where $x_{i}$ is the prediction for $ith$ medical concept  ($i \in \left \{ 0,\; 1, \; \cdots, \; K \right \}$), and $y_{i}$ is the ground-truth label for $ith$ medical concept where $y_i \in \left\{0, \; 1\right\}$, with $y_i = 1$ meaning the input image has the corresponding medical concept while $y_i = 0$ meaning the opposite. \\




\subsection{Report Decoder}~\label{sec:2-4}
The report decoder aims to generate the output report conditioned on the attended visual embeddings $\mathbf{V}_i = {\hat{\mathbf{Q}}^(m)}_{m=0}^M$ from image encoder and the medical concepts embeddings $\mathbf{V}_c$ from MCGN. Therefore, we first fused those embeddings by the equation: $\mathbf{v}_f = \mathbf{W}_f[\hat{\mathbf{Q}}^{(0)}, \hat{\mathbf{Q}}^{(1)}, \cdots , \hat{\mathbf{Q}}^{(M)}] + \mathbf{V}_c$, where $\mathbf{W}_f$ is a learnable parameter. The input of transformer decoder is thus set as the concatenation of the fused feature $\mathbf{v}_f$, the regional features $\mathbf{K}^{(M)}$ and $\mathbf{V}^{(M)}$ output by image encoder, and the input word $\mathbf{w}_t$. Generally, the first decoder sub-layer takes $\mathbf{v}_f$ as the query to calculate sparse nonlinear attention with the word embeddings (taken as keys and values) of $\mathbf{w}_t$ to obtain the visual-enhanced word embeddings. Then the output of the first sub-layer will be coupled with the image encoder's output $\mathbf{K}^{(M)}$ and $\mathbf{V}^{(M)}$ as $\mathbf{Q}$, $\mathbf{K}$, $\mathbf{V}$ input to another sparse attention block and repeat this process. Note that we employ residual connections around each sub-layers similar to the encoder, followed by layer normalization. The decoder is also composed of a stack of N = 6 identical layers.

We train our model parameters $\theta$ by minimizing the negative log-likelihood of $\mathbf{P}(T)$ given the image features: 
\begin{equation}
    {\mathcal L}_{CE} = -\sum_{i=1}^{N}log P_{\theta}({\mathbf t}_i|{\mathbf I}, {\mathbf t}_{i-1}, \cdots, {\mathbf t}_1)
\end{equation}
where $\mathbf{P}({\mathbf t}_i|{\mathbf I}, {\mathbf t}_{i-1}, \cdots, {\mathbf t}_1)$ represents the probability predicted by the model for the  $i$-th word based on the information of the image $\mathbf{I}$ and the first $(i-1)$ words.\\

\noindent\textbf{Overall objective function}~~~Our overall objective integrates the two losses regarding report generation and multi-class classification, which is defined as: 
\begin{equation}
    {\mathcal L}_{all} = \lambda_{CE}{\mathcal L}_{CE} + \lambda_{MLC}{\mathcal L}_{MLC}
\label{equ:total_loss}
\end{equation}
The hyper-parameters $\lambda_{CE}$ and $\lambda_{MLC}$ balance the two losses terms, and their values are given in Section.~\ref{Sec:Experiments}.

\section{Experiments}\label{Sec:Experiments}

\subsection{Data Collection}~\label{sec:3-1}
MIMIC-CXR~\cite{2019MIMIC} is the recently released largest dataset to date containing both chest radiographs and free-text reports. It consists of 377110 chest x-ray images and 227835 reports from 64588 patients of the Beth Israel Deaconess Medical Center. In our experiment, we adopt MIMIC-CXR's official split following~\cite{chen2020generating,CVPR201_PPKD} for a fair comparison, resulting in a total of 222758 samples for training, and 1808 and 3269 samples for validation and test, respectively. We convert all tokens to lowercase characters and remove non-word illegal characters and the words whose occurring frequency is lower than 5, counting to 5412 unique words remaining in the dataset.
\begin{table}[h]
\caption{Performance comparison on MIMIC-CXR dataset.}\label{tab:1}
\begin{tabular}{cccccccc}
\hline
Methods                       & Bleu-1                       & Bleu-2         & Bleu-3         & Bleu-4         & ROUGE                        & METEOR                       & CIDEr                        \\ \hline
Show-Tell~\cite{vinyals2015showandtell}                     & 0.308                        & 0.190          & 0.125          & 0.088          & 0.256                        & 0.122                        & 0.096                        \\
Att2in~\cite{xu2016showattandtell}                        & 0.314                        & 0.198          & 0.133          & 0.095          & 0.264                        & 0.122                        & 0.106                        \\
AdaAtt~\cite{2017Knowing}                        & 0.314                        & 0.198          & 0.132          & 0.094          & 0.267                        & 0.128                        & 0.131                        \\
Transformer~\cite{vaswani2017attentionisallyouneed}                   & 0.316                        & 0.199          & 0.140          & 0.092          & 0.267                        & 0.129                        & 0.134                        \\
M2Transformer~\cite{2020Meshed}                 & 0.332                        & 0.210          & 0.142          & 0.101          & 0.264                        & 0.134                        & 0.142                        \\
R2Gen~\cite{chen2020generating}                         & 0.353                        & 0.218          & 0.145          & 0.103          & 0.277                        & 0.142                        & 0.141                        \\
R2GenCMN~\cite{chen2021cross}                      & 0.353                        & 0.218          & 0.148          & 0.106          & 0.278                        & 0.142                        & 0.143                        \\
PPKED$^{\dagger}$~\cite{CVPR201_PPKD}                    & 0.360                        & 0.224          & 0.149          & 0.106          & 0.284                        & 0.149                        & 0.237                        \\
Self-boost~\cite{wang2021self}                    & 0.359                        & 0.224          & 0.150          & 0.109          & 0.277                        & 0.141                        & 0.270                        \\ \hline
baseline                          &  0.352                        & 0.215          & 0.145          & 0.105          & 0.264                        & 0.133                        & 0.240                        \\
baseline+MSA                           & 0.369                        & 0.231          & 0.159          & 0.117          & 0.280                        & 0.142                        & 0.297                        \\
Ours(baseline+MSA+MCGN)                          & 0.373                        & 0.235          & 0.162          & 0.120          & 0.282                        & 0.143                        & 0.299                        \\
Ours+RL                     & \textbf{0.413}               & \textbf{0.266} & \textbf{0.186} & \textbf{0.136} & \textbf{0.298}               & \textbf{0.170}               & \textbf{0.429}               \\ \hline
\end{tabular}
\end{table}
\vspace{-5mm}
\subsection{Experimental Settings}
\noindent \textbf{Evaluation Metrics}~~~Following the standard evaluation protocol\footnote{https://github.com/tylin/coco-caption}, we utilise the most widely used  BLEU scores~\cite{Kishore2002bleu}, ROUGE-L~\cite{lin-2004-rouge}, METEOR~\cite{banerjee-lavie-2005-meteor} and CIDER~\cite{vedantam2015cider} as the metrics to evaluate the quality of the generated text report.

\noindent \textbf{Implementation Details}~~~We extract our pre-computed image features by a pre-trained "CLIP-B/16"~\cite{clip}~\footnote{https://github.com/openai/CLIP/} model, resulting in a regional feature matrix $f \in \mathbb{R}^{768 \times 196}$ (reshaped from $768 \times 14 \times 14$). The number of memory vectors is set to 3. The number of the layers in both the image transformer encoder and the text transformer decoder is set to $6$ and the number of heads in multi-head attention is set to $8$. The hyper-parameters $\lambda_{CE}$, $\lambda_{MLC}$ are set as $1$ and $5$, respectively. We train our model using Adam optimizer~\cite{adam} on eight NVIDIA TESLA V100 with a mini-batch size of $32$ for each GPU. The learning rates is set to be $5e-5$, and the model is trained in a total of $60$ epochs. At the inference stage, we adopt the beam search strategy and set the beam size as 3. In addition, following~\cite{rennie2017selfcritical}, we also apply reinforcement learning with CIDEr as the reward and further train the model for 20 epochs to boost the performance. 

\begin{figure}[h]
\includegraphics[width=\textwidth]{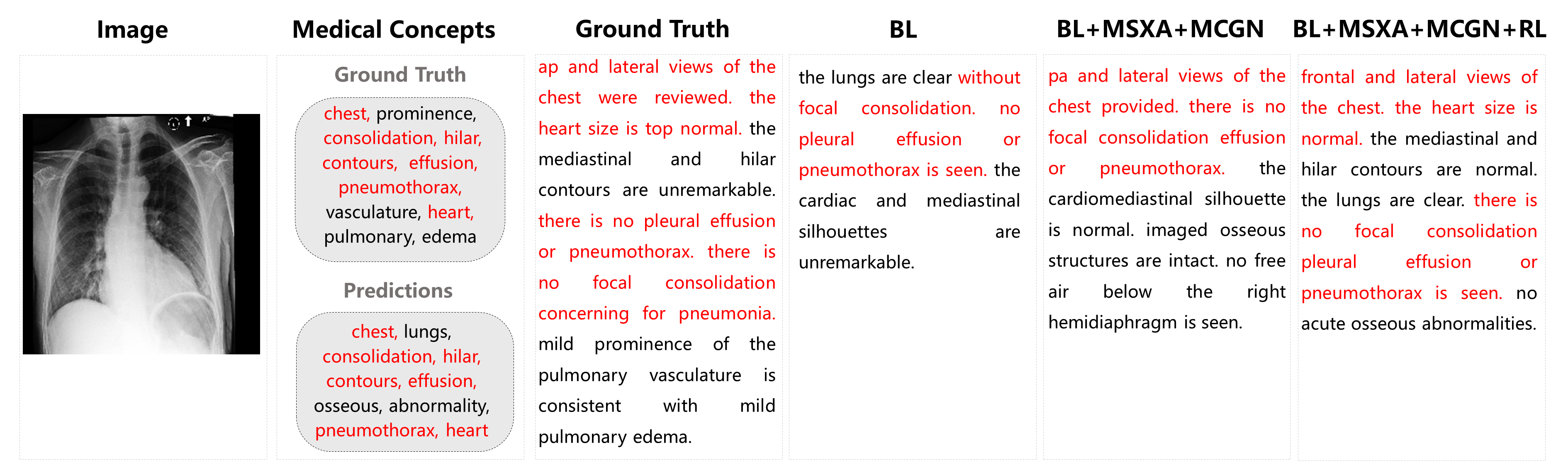}
\caption{An example report generated by the proposed model. The correctly predicted medical concepts and the key information in the report are marked red.}
\label{fig:2}
\end{figure}
\vspace{-7mm}

\subsection{Results and Discussion}
\noindent\textbf{Comparison with SOTA} Table~\ref{tab:1} summarizes the performance comparisons between the state-of-the-art methods and our proposed model on the MIMIC-CXR Official test split. Specifically, there are five SOTA image captioning methods in the comparison, including Show-tell~\cite{vinyals2015showandtell}, AdaAtt~\cite{2017Knowing}, Att2in~\cite{topdown}, Transformer~\cite{vaswani2017attentionisallyouneed}, and M2transformer~\cite{2020Meshed}. Moreover, we also compare with four SOTA medical report generation methods: including R2Gen~\cite{chen2020generating}, R2GenCMN~\cite{chen2021cross}, PPKED~\cite{CVPR201_PPKD} and  Self-boost~\cite{wang2021self}. For PPKED~\cite{CVPR201_PPKD}, we quote the performance from their paper (marked with $\dagger$ in Table.~\ref{tab:1}) since this model does not release the code. For the other methods in comparison, we download the codes released publicly and re-run them on the MIMIC-CXR dataset with the same experimental setting as ours, so they are strictly comparable.

As shown in Table~\ref{tab:1}, our proposed method is the best performer over almost all evaluation metrics among the comparing methods, even without the reinforcement learning. Specifically, the two very recent medical report generation models, R2Gen~\cite{chen2020generating} and Self-boost~\cite{wang2021self}, perform better than other image captioning methods but still lose to ours. R2Gen adopts a memory-augmented transformer decoder, but its image encoder still relies on the CNN model, which may fail to identify fine-grained differences of radiographic images. For the Self-boost model, despite utilizing an image-text matching network to help the image encoder learn fine-grained visual differences, it builds upon CNN and LSTM and does not incorporate semantic information into the report generation process like ours. It is also noted that although performing slightly inferior than PPKED~\cite{CVPR201_PPKD} on Meteor and Rouge, our models shows clear advantages over PPKED on Bleu and CIDEr metrics. PPKED is a transformer based model and encodes semantic concepts into an external knowledge graph to guide the report generation. However, this knowledge graph is sparse with only 25 nodes (concepts) and requires extra efforts to construct, while our fine-grained medical concepts can be easily picked up from the reports.



\noindent\textbf{Ablation Study} Table.~\ref{tab:1}(lower part) shows the results of the ablation study to single out the contributions of each component of our model. We remove MCGN from our model and replace MSA with conventional self-attention~\cite{vaswani2017attentionisallyouneed} as the baseline to verify the performance improvements brought by the proposed attention and our concepts generation network. In table~\ref{tab:1}, there are three components: MSA, MCGN, and RL, representing Memory-augmented Sparse Attention, Medical Concepts Generation Network, and Reinforcement Learning, respectively. The symbols ``+" or ``-" indicate the inclusion or exclusion of the following component. The benefit of using MSA block can be well reflected by the improvement from ``baseline" to  ``baseline+MSA". As shown, the performance can be further boosted by additionally introducing medical concepts (``baseline+MSA+MCGN").  Moreover, as mentioned, even removing reinforcement learning, our model still significantly outperforms the comparing methods in Table~\ref{tab:1}. Our proposed model produces more accurate and descriptive findings compared with baseline, as the qualitative results show in Figure.~\ref{fig:2}.

\vspace{-2mm}
\section{Conclusions}
We propose a medical report generation model utilizing sparse nonlinear attention in the transformer-structured encoder-decoder paradigm. Compared with prior arts, our proposed model considers higher-order interactions across the input feature vectors and thus can better cater for the fine-grained radiographic images. Our proposed Medical Concepts Generation Network introduces rich semantic concepts and encodes them as semantic information to benefit the radiographic report generation task. Extensive experimental results demonstrate that our approach surpasses the state-of-the-art with a large margin.

\vspace{5mm}
\noindent\textbf{Acknowledgment} Prof Xiu Li was supported by the National Key R $\&$ D Program of China (No.2020AAA0108303) and NSFC 41876098.

%
%
%
\bibliographystyle{splncs04}
\bibliography{paper1374}
\end{document}